\documentclass[conference,a4paper]{IEEEtran}
\IEEEoverridecommandlockouts
\usepackage[nocompress]{cite}
\usepackage{amsmath,amssymb,amsfonts}
\usepackage{graphicx}
\usepackage[font=footnotesize]{caption}
\usepackage{textcomp}
\usepackage{xcolor}
\usepackage{url}
\usepackage{booktabs}
\usepackage{balance}
\usepackage{microtype}
\usepackage[utf8]{inputenc}
\usepackage{geometry}
\usepackage[hidelinks]{hyperref}
\hypersetup{pdftitle={When Is Personalized Federated Learning Worth Its Deployment Cost?},pdfauthor={Beyazit Bestami Yueksel, Ayse Yilmazer Metin, Emrah Dikbiyik},pdfkeywords={federated learning, personalization, utility, deployment cost, allocation regime},pdfcreator={pdfTeX},pdfproducer={LaTeX with hyperref}}

\renewcommand{\refname}{References}
\renewcommand{\figurename}{Fig.}
\renewcommand{\tablename}{TABLE}
\renewcommand{\abstractname}{Abstract}

\begin{document}
\title{When Is Personalized Federated Learning Worth Its Deployment Cost?}

\makeatletter
\def\@IEEEauthorblockNfont{\fontsize{11pt}{13pt}\selectfont}
\def\@IEEEauthorblockAfont{\fontsize{10pt}{12pt}\selectfont}
\makeatother

\author{\IEEEauthorblockN{Beyazıt Bestami Yüksel}
\IEEEauthorblockA{\textit{Computer Engineering} \\
\textit{Istanbul Technical University}\\
Istanbul, Türkiye \\
yukselbe18@itu.edu.tr}
\and
\IEEEauthorblockN{Ayşe Yılmazer Metin}
\IEEEauthorblockA{\textit{Computer Engineering} \\
\textit{Istanbul Technical University}\\
Istanbul, Türkiye \\
yilmazerayse@itu.edu.tr}
\and
\IEEEauthorblockN{Emrah Dikbıyık}
\IEEEauthorblockA{\textit{Computer Engineering} \\
\textit{Marmara University}\\
Istanbul, Türkiye \\
emrah.dikbiyik@marmara.edu.tr}}

\maketitle

\begin{abstract}
Personalized federated learning (PFL) may improve client-level performance, but deeper personalization increases deployment cost. We formulate personalization-depth selection as a utility-based empirical decision problem in which client utility equals Macro-F1 minus a cost-sensitivity weight times deployment cost, over six personalization depths labeled P0 through P5. We compare fixed policies with an offline per-client Oracle reference and report allocation maps over Dirichlet data heterogeneity and cost sensitivity. Although evaluated on this hierarchy, the formulation can accommodate any finite candidate set. On CIFAR-10, CIFAR-100, and a PTB-XL subset, personalization shows empirical, setting-dependent decision regimes rather than a universal benefit: higher cost sensitivity shifts Oracle mass toward the global-only depth P0; low cost sensitivity can favor deeper actions. Patterns are grid observations (three seeds per cell), not theoretical boundaries. Best-uniform is competitive with Oracle assignment (6.5 percent plus or minus 7.8 percent relative regret; 95 percent confidence interval 5.9 to 7.1), whereas random and personalize-all incur much larger regret (189.3 percent plus or minus 367.1 percent and 409.0 percent plus or minus 803.2 percent). Under the same utility, Ditto full-model personalization is substantially suboptimal once cost is counted. The contribution is an offline characterization of when personalization depth is worth its cost, not a new PFL optimizer or deployment-time scheduler.
\end{abstract}

\renewcommand{\IEEEkeywordsname}{Keywords}

\section{Introduction}
Federated learning (FL) trains models across decentralized clients without pooling raw data. Personalized FL (PFL) adapts models to client-specific distributions through fine-tuning, regularization, representation sharing, or client-specific layers~\cite{fallah2020perfedavg,li2021ditto}. Most PFL work therefore asks \emph{how} to personalize.

Deployment raises a complementary question: \emph{when} is personalization worth its deployment cost? Personalization consumes computation, parameters, memory, communication, and operational complexity. Personalizing every client may waste resources when the global model already suffices; a global-only policy may fail for strongly heterogeneous clients. We present a utility-based empirical decision framework for personalization-depth selection: measure predictive performance and deployment cost, form a performance--cost utility, obtain an offline Oracle reference by ex-post argmax over a finite action set, and characterize regret of practical uniform policies. Experiments instantiate $\mathcal{A}=\{P_0,\ldots,P_5\}$ and compare global-only, best-uniform, random, personalize-all, and Ditto against that Oracle. This is not a new PFL optimizer, not a deployment-time scheduler, and not a universal theory of personalization.

Contributions: (i)~we formulate personalization-depth selection as a finite performance--cost utility-maximization problem; (ii)~we instantiate the decision space with the six-level $P_0$--$P_5$ hierarchy and characterize offline Oracle assignments across heterogeneity and cost-sensitivity regimes; (iii)~we quantify regret of deployable uniform policies relative to the Oracle reference and identify empirical decision regimes in which deeper personalization is or is not worth its cost; and (iv)~we note that the formulation is action-space agnostic while limiting empirical claims to the evaluated $P_0$--$P_5$ setting (including a PTB-XL subset and Ditto under the same utility).

\section{Related Work}
\subsection{Classical Federated Learning}
FedAvg~\cite{mcmahan2017fedavg} learns a shared global model. FedProx~\cite{li2020fedprox}, SCAFFOLD~\cite{karimireddy2020scaffold}, FedNova~\cite{wang2020fednova}, and adaptive server optimization~\cite{reddi2021adaptive} address heterogeneity. Surveys document open problems and scale~\cite{kairouz2021advances,bonawitz2019towards}. Non-IID effects have been measured directly~\cite{hsieh2020noniid,zhao2018noniid}.

\subsection{Personalized FL: \emph{How} to Personalize}
The PerFedAvg method~\cite{fallah2020perfedavg} uses meta-learning; Ditto~\cite{li2021ditto} and pFedMe~\cite{dinh2020pfedme} regularize personalized objectives. FedPer~\cite{arivazhagan2019fedper} personalizes classifier heads; FedRep~\cite{collins2021fedrep} separates representation and classifier. FedBN~\cite{li2021fedbn} uses local batch normalization for feature shift; FedBABU~\cite{oh2022fedbabu} decouples body and head training; APFL~\cite{deng2020apfl} adaptively mixes global and local models; and FedAMP~\cite{huang2021fedamp} clusters similar clients. Adapters appear in pFedLoRA~\cite{zhang2023pfedlora}, following LoRA~\cite{hu2021lora}. FedALA~\cite{zhang2023fedala} and personalization-layer decoupling~\cite{zhu2024fedtsd} refine client capacity. Peer-reviewed 2025 methods continue this axis: FedFSA~\cite{xing2025fedfsa} and PFedCS~\cite{wu2025pfedcs}. Large-model personalization includes FLoRA~\cite{wang2024flora} and OpenFedLLM~\cite{ye2024openfedllm}. Surveys catalogue many ``how'' methods~\cite{tan2022surveypfl}. Deployment cost rarely enters as an explicit shared utility over personalization depth.

\subsection{This Work: \emph{When} Personalization Is Worth Its Cost}
We ask under which $(\text{task},\alpha,\lambda)$ client $i$ should personalize, and to what depth. Rather than proposing another PFL trainer or online allocator, we evaluate fixed policies, an offline Oracle reference, and Ditto under~\eqref{eq:utility}, map empirical decision regimes over $(\alpha,\lambda)$, and report regret with confidence intervals.

\subsection{Resource-Aware FL and Benchmarks}
Resource-aware FL uses local SGD~\cite{stich2018localsgd}, quantization~\cite{alistarh2017qsgd}, compression~\cite{konevcny2016federated}, and client selection such as FedCS~\cite{nishio2019client}, Oort~\cite{lai2021oort}, and TiFL~\cite{cho2020tifl}. HeteroFL~\cite{diao2021heterofl} trains heterogeneous local capacities toward one global model; related personalized communication-efficient designs appear in industrial IoT~\cite{wang2023personalized}. Hardware-aware FL further couples client selection and local workload to device profiles under communication and latency constraints~\cite{yuksel2026hardwarefl}. Benchmarks include LEAF~\cite{caldas2018leaf}, FedScale~\cite{lai2022fedscale}, FedML~\cite{he2020fedml}, FLamby~\cite{ogierdu2022flamby}, and PTB-XL~\cite{wagner2020ptbxl}; multi-institutional ECG FL has also been studied under encryption-based privacy constraints~\cite{yuksel2026ecgfl}.

\section{Formulation and Experimental Design}
Clients $i\in\{1,\ldots,N\}$ participate in FedAvg~\cite{mcmahan2017fedavg} to obtain a global model, then apply a client-specific personalization action on held-out data.

\paragraph{Decision Formulation.}
We formulate personalization-depth selection as a finite utility-maximization problem. The decision maker is a system designer, deployment planner, or FL practitioner selecting a personalization strategy under a specified performance--cost preference. The decision variable is an action $a\in\mathcal{A}$. In our experiments the finite action space is the six-level personalization hierarchy adopted in this study,
\begin{equation}
\mathcal{A}=\{P_0,P_1,P_2,P_3,P_4,P_5\},
\label{eq:action_space}
\end{equation}
(Table~\ref{tab:ladder}): an ordered operationalization of personalization depth for controlled analysis, not a universal taxonomy or the complete space of PFL strategies. For client $i$, utility is
\begin{equation}
U_i(a)=\mathrm{Macro\text{-}F1}_i(a)-\lambda\,\mathrm{Cost}_i(a),
\label{eq:utility}
\end{equation}
where $\mathrm{Macro\text{-}F1}_i(a)$ is predictive performance, $\mathrm{Cost}_i(a)\in[0,1]$ is the normalized deployment-cost proxy (Section~\ref{sec:cost}), and $\lambda\geq0$ is the cost-sensitivity preference. The preferred action is
\begin{equation}
a_i^{\star}=\arg\max_{a\in\mathcal{A}} U_i(a).
\label{eq:oracle}
\end{equation}
Scope conditions only: $a$ must belong to the evaluated set $\mathcal{A}$; performance and cost must be measurable or estimable; the cost model and $\lambda$ are specified before utility comparison; and the Oracle reference is restricted to evaluated candidates. This specifies the decision problem and its empirical reference solution; it does not constitute a deployment-time policy.

The linear utility is intentionally a transparent first-order approximation. It provides an interpretable performance--cost trade-off through a single parameter $\lambda$ under a constant marginal cost penalty, while avoiding deployment-specific nonlinear assumptions that cannot be justified uniformly across heterogeneous platforms. Nonlinear, thresholded, or hardware-specific utilities may be appropriate elsewhere.

\paragraph{Role of the Oracle.}
The Oracle is an offline, ex-post reference rule that selects the utility-maximizing action after all candidate actions have been evaluated~\eqref{eq:oracle}. It is an empirical upper-bound reference over measured utilities in $\mathcal{A}$: it is not deployable, not predictive, does not generate utility values, does not possess future information, does not avoid exhaustive evaluation, and is not an online scheduler. Methodological pipeline: measure predictive performance; estimate deployment cost; compute $U_i(a)$; select $a_i^{\star}$ with the offline Oracle; compute regret $r_i(\pi)=U_i(a_i^{\star})-U_i(\pi_i)$ as a downstream evaluation measure; then analyze the decision landscape. We report mean regret, relative regret $\bar{r}_{\mathrm{rel}}=\bar{r}/\max(|\bar{U}(\textsc{Oracle})|,\epsilon)$, and Oracle win rate over $(\alpha,\lambda)$~\cite{hsieh2020noniid} via P0 ratio, deep ratio ($P_3$--$P_5$), and gaps to fixed baselines.

\paragraph{Generality of the Oracle Framework.}
Although instantiated with $P_0$--$P_5$, the formulation is agnostic to the semantic meaning of candidate actions. More generally, $\mathcal{A}$ may contain any finite set of personalization or deployment alternatives---adapter configurations, LoRA ranks, alternative PFL methods, model-compression choices, or edge--cloud execution modes. The offline Oracle requires only measured utilities and selects their argmax. The generality is methodological rather than empirical: the present experiments validate the formulation only for the $P_0$--$P_5$ action space considered in this work.

\begin{table}[t]
\caption{PERSONALIZATION HIERARCHY ADOPTED IN THIS STUDY (EXPERIMENTAL $\mathcal{A}$).}
\label{tab:ladder}
\centering
\footnotesize
\begin{tabular}{ll}
\toprule
\textbf{Action} & \textbf{Locally trainable components}\\
\midrule
$P_0$ & None; global-only model\\
$P_1$ & Classifier head\\
$P_2$ & Adapter and head\\
$P_3$ & Last encoder block, adapter, head\\
$P_4$ & Last two encoder blocks, adapter, head\\
$P_5$ & Full model personalization\\
\bottomrule
\end{tabular}
\end{table}

\paragraph{Why this hierarchy?}
$\mathcal{A}$ is a practical discretization spanning global-only inference to full local fine-tuning by progressively increasing trainable client-specific components; it does not exhaust every PFL design. Cost generally increases with depth by construction; predictive performance and utility need not increase monotonically. $P_0$ is FedAvg~\cite{mcmahan2017fedavg}; $P_1$ mirrors head-only personalization~\cite{arivazhagan2019fedper}; $P_2$ adds an adapter plus head~\cite{hu2021lora,zhang2023pfedlora}; $P_3$--$P_4$ unfreeze last encoder block(s)~\cite{collins2021fedrep}; $P_5$ is full-model fine-tuning (Ditto cost extremity~\cite{li2021ditto}). Six ordered depths resolve empirical regime-change patterns without enumerating every layer mask.

\paragraph{Datasets and protocol.}
Experiments use CIFAR-10, CIFAR-100, an auxiliary CIFAR-10 vision split, and a PTB-XL subset for binary ECG classification~\cite{wagner2020ptbxl}. The subset uses Lead~II only, the WFDB \texttt{records100} release at 100\,Hz, at most 5000 records, and a binary Normal vs.\ Abnormal label derived from SCP-ECG codes. Unlike privacy-preserving multi-institutional ECG FL with homomorphic encryption~\cite{yuksel2026ecgfl}, our ECG evaluation is a cost-aware personalization-depth study on a public PTB-XL subset rather than a clinical deployment system.

We run FedAvg for 5 global rounds and 3 local epochs, then evaluate all actions in $\mathcal{A}$ on held-out data. Client-level metrics are computed for each $(\text{dataset},\text{seed},\alpha,N)$ configuration; $\lambda$ sweeps are applied post hoc. Allocation map: 378 cells (3 datasets $\times$ 7 $\alpha$ $\times$ 6 $\lambda$ $\times$ 3 seeds), $\alpha\in\{0.01,0.05,0.1,0.3,0.5,1.0,10.0\}$, $\lambda\in\{0,0.1,0.3,0.5,1.0,2.0\}$, 20 clients. Sparsity validation: 54 cells (3 datasets $\times$ 3 $\alpha$ $\times$ 3 seeds $\times$ 2 client counts), $\alpha\in\{0.1,0.5,1.0\}$. Regret study: 594 cells (99 configurations $\times$ 6 $\lambda$ $\times$ 5 policies). Ditto~\cite{li2021ditto} is scored with P5-equivalent cost. Image tasks use a ResNet-style encoder; ECG uses a 1D ResNet on Lead~II (1000 samples, per-lead z-score). Dirichlet client splits use 70/15/15 train/val/test.

\paragraph{Baselines.}
The \emph{global-only} policy uses $P_0$ for all clients; the \emph{personalize-all} policy uses $P_5$ for all; the \emph{random} policy samples depths without utility awareness; and the \emph{best-uniform} policy chooses the best single depth for all clients within a configuration. The best-uniform policy is intentionally strong because it represents a realistic low-complexity deployment alternative to heterogeneous per-client allocation.

\subsection{Cost Model and Post-Hoc Sweeps}
\label{sec:cost}
Five proxies enter a weighted raw score (Table~\ref{tab:cost_components}), then per-client min--max over $\{P_0,\ldots,P_5\}$. Let $R_i^{\min}{=}\min_{a'}\mathrm{RawCost}_i(a')$, $R_i^{\max}{=}\max_{a'}\mathrm{RawCost}_i(a')$, and $\varepsilon{=}10^{-12}$ be a numerical-stability constant that does not alter reported results when $R_i^{\max}{>}R_i^{\min}$:
\begin{equation}
\mathrm{Cost}_i(a)=\frac{\mathrm{RawCost}_i(a)-R_i^{\min}}{R_i^{\max}-R_i^{\min}+\varepsilon}.
\label{eq:cost_norm}
\end{equation}
When $R_i^{\max}-R_i^{\min}<\varepsilon$, all normalized costs are set to zero.
\paragraph{Why normalize within each client across the six actions?}
Absolute proxy magnitudes differ across clients (local data size, measured wall time). Within-client min--max places each client's six depths on a shared $[0,1]$ scale so $\lambda$ has a comparable exchange-rate meaning: the cheapest depth has cost $0$ and the most expensive has cost $1$. Without this step, $\lambda$ would confound absolute scale differences with the accuracy--cost trade-off under study.

Let $\mathrm{Par}_i{=}\mathrm{Params}_i/10^5$, $\mathrm{Fl}_i{=}\mathrm{FLOPs}_i/10^8$, $\mathrm{Tm}_i{=}\mathrm{Time}_i$, $\mathrm{Cc}_i{=}\mathrm{Comm}_i/10^6$, $\mathrm{Mm}_i{=}\mathrm{Mem}_i/100$. Default weights $(w_{\mathrm{p}},w_{\mathrm{f}},w_{\mathrm{t}},w_{\mathrm{c}},w_{\mathrm{m}}){=}(0.25,0.20,0.25,0.15,0.15)$ match the implementation:
\begin{equation}
\begin{aligned}
\mathrm{RawCost}_i(a)=m_i\bigl(&w_{\mathrm{p}}\mathrm{Par}_i(a)+w_{\mathrm{f}}\mathrm{Fl}_i(a)+w_{\mathrm{t}}\mathrm{Tm}_i(a)\\
&+w_{\mathrm{c}}\mathrm{Cc}_i(a)+w_{\mathrm{m}}\mathrm{Mm}_i(a)\bigr).
\end{aligned}
\label{eq:raw_cost}
\end{equation}
Multiplier $m_i$ comes from simulated LOW/MEDIUM/HIGH profiles (lower capability $\Rightarrow$ larger $m_i$; approximately $4.21/1.87/1.00$). Because $m_i$ is constant across actions for a client, it cancels in~\eqref{eq:cost_norm}: hardware tiers affect raw absolute scale but not normalized $\mathrm{Cost}_i(a)$ under the present protocol. The same component definitions, pre-weight scales, default weights, and within-client min--max normalization are used across datasets and models. Reported allocation maps are therefore driven by relative action costs under the chosen weight vector; device-specific hardware calibration remains future work. This is a relative deployment-cost proxy, not measured device energy or monetary cost.

\paragraph{Weights and conditionality.}
Components cover footprint, compute, latency, communication, and memory~\cite{kairouz2021advances,konevcny2016federated}. Pre-weight scalings bring units onto comparable numeric ranges before weighting. Alternative weight vectors may alter relative action spacing, action-cost ordering, Oracle assignments, and allocation maps; we did not sweep weights experimentally. All reported findings are conditional on the default weighting above.

\begin{table}[t]
\caption{COST COMPONENTS, PRE-WEIGHT SCALING (APPLIED BEFORE WEIGHTING), AND DEFAULT WEIGHTS.}
\label{tab:cost_components}
\centering
\footnotesize
\resizebox{\columnwidth}{!}{%
\begin{tabular}{@{}llll@{}}
\toprule
\textbf{Comp.} & \textbf{Meaning} & \textbf{Pre-weight scale} & \textbf{Weight}\\
\midrule
Params & Trainable parameters at $a$ & $\div10^5$ & 0.25\\
FLOPs & Local train+eval estimate & $\div10^8$ & 0.20\\
Time & Local training wall time (s) & --- & 0.25\\
Comm & Model-delta bytes & $\div10^6$ & 0.15\\
Mem & Param+activation proxy (MB) & $\div100$ & 0.15\\
\bottomrule
\end{tabular}%
}
\end{table}

\paragraph{Interpretation of $\lambda$.}
Parameter $\lambda$ encodes a relative operational preference over the normalized cost scale, not a universal monetary unit. Qualitatively, low $\lambda$ is performance-oriented, medium $\lambda$ is balanced, and high $\lambda$ is resource-sensitive. Table~\ref{tab:lambda} illustrates our sweep---not calibrated mappings to joules, battery drain, measured latency, monetary cost, or SLA penalties.

\begin{table}[t]
\caption{QUALITATIVE $\lambda$ READINGS (NOT HARDWARE-CALIBRATED).}
\label{tab:lambda}
\centering
\footnotesize
\resizebox{\columnwidth}{!}{%
\begin{tabular}{@{}cll@{}}
\toprule
$\boldsymbol{\lambda}$ & \textbf{Qualitative reading} & \textbf{Example regime (illustrative only)}\\
\midrule
0.0 & Accuracy-only & Ignore cost; Macro-F1 maximization\\
0.1 & Mild cost preference & Quality dominates; cost secondary\\
0.3 & Balanced default & Default in sparsity tables\\
0.5--1.0 & Strong cost preference & Edge / battery-constrained settings\\
2.0 & Cost-dominant & Cost dominates small F1 gains\\
\bottomrule
\end{tabular}%
}
\end{table}

\subsection{Evaluation Protocol}
All policies share the same client partitions and held-out splits. After FedAvg, each personalization action in $\mathcal{A}$ is evaluated locally. Regret against the best-uniform policy measures how much utility practical uniform selection loses relative to client-specific offline Oracle assignments. We report mean$\pm$std and 95\% bootstrap confidence intervals (10{,}000 resamples) over cells; Student's $t$ and bootstrap intervals agree closely for best-uniform regret ($t$-CI $[5.86,7.12]$, bootstrap $[5.89,7.15]$). Paired Wilcoxon tests (Table~\ref{tab:sig}) are descriptive (no family-wise correction).

\section{Results}
\subsection{Finding 1: Empirical regime-change patterns}
\begin{figure*}[t]
\centering
\includegraphics[width=\textwidth,height=0.70\textheight,keepaspectratio]{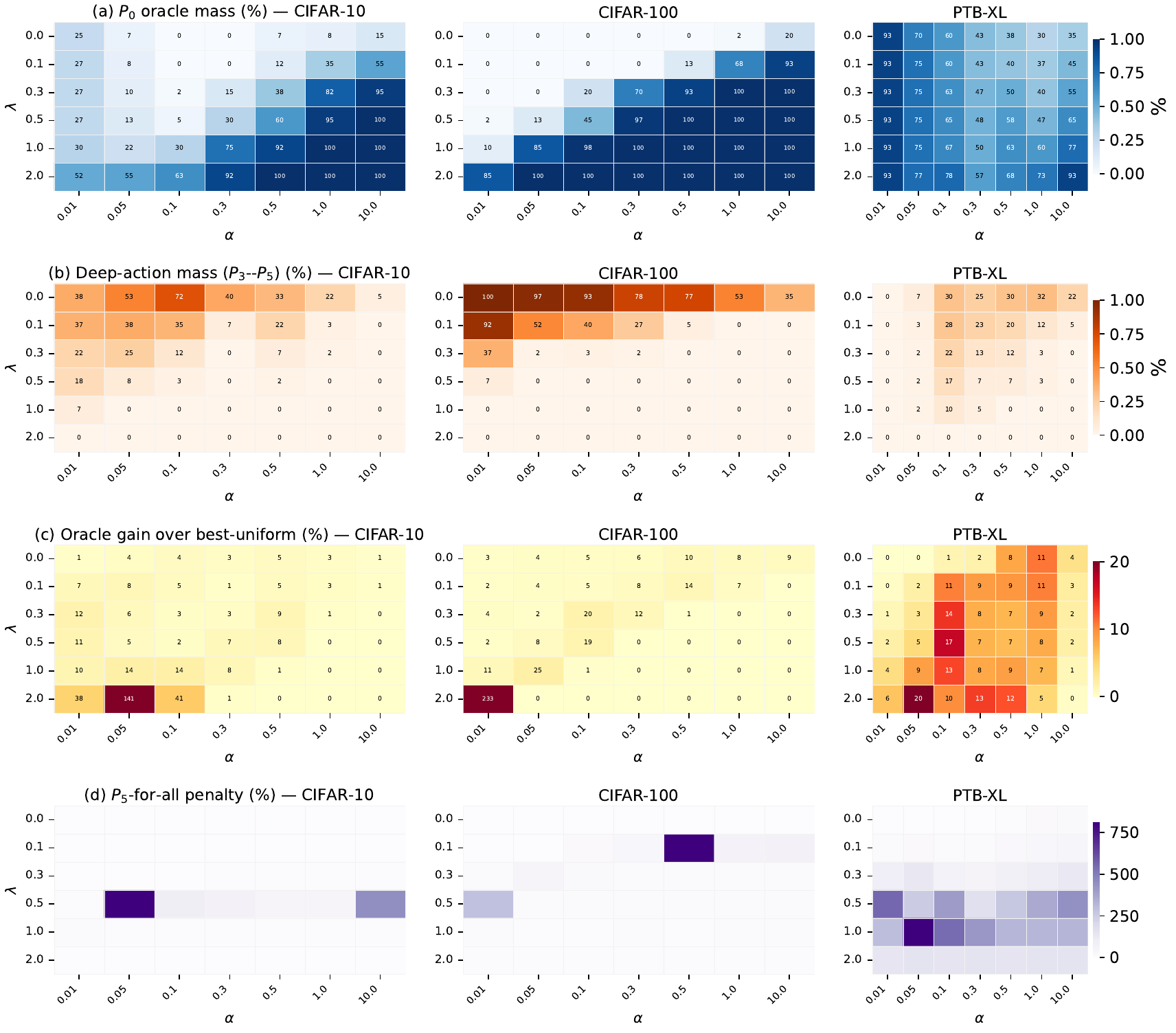}
\caption{Empirical offline-Oracle allocation maps over Dirichlet heterogeneity $\alpha$ and cost sensitivity $\lambda$. Each row shows CIFAR-10, CIFAR-100, and PTB-XL for one metric: (a)~$P_0$ mass (\%), (b)~deep-action mass $P_3$--$P_5$ (\%), (c)~Oracle gain over the best-uniform policy (\%), (d)~$P_5$-for-all penalty (color scale; cell labels omitted due to extreme magnitude range). The maps show regime-dependent allocation patterns on the finite experimental grid under $\mathcal{A}=\{P_0,\ldots,P_5\}$.}
\label{fig:phase}
\end{figure*}

Fig.~\ref{fig:phase} shows that offline Oracle assignment changes sharply over the $(\alpha,\lambda)$ grid. As $\lambda$ increases, mass shifts toward $P_0$ because marginal Macro-F1 gain no longer offsets deployment cost; at low $\lambda$, deeper actions $P_3$--$P_5$ reappear where client data are sufficiently heterogeneous. We describe these as empirical regime-change patterns on a finite three-seed grid---not theoretically derived phase transitions. Observed boundaries depend on datasets, architectures, seeds, heterogeneity, cost normalization, $\lambda$, and the $P_0$--$P_5$ action space.

The pattern appears across modalities but occupies different heatmap regions. On CIFAR-10, high cost sensitivity can move the Oracle almost entirely to $P_0$ (P0 ratio 100\% at $\alpha{=}0.5,\lambda{=}2.0$), whereas low-cost, high-heterogeneity regimes still support deep personalization (deep ratio 53.3\% at $\alpha{=}0.05,\lambda{=}0.0$). CIFAR-100 concentrates on shallow/global actions in sparsity validation, with P0 ratio $0.620\pm0.383$ at $\lambda{=}0.3$ (CI $[0.44,0.78]$) and near-zero deep ratio. The PTB-XL subset occupies a different regime: at $\lambda{=}0.3$, P0 ratio is $0.552\pm0.167$ (CI $[0.481,0.631]$) and deep ratio is $0.127\pm0.110$ (CI $[0.079,0.178]$).

These maps are empirical design guidance under the evaluated settings and $P_0$--$P_5$ hierarchy---not universal deployment rules.

\subsection{Finding 2: The best-uniform policy is a strong baseline}
Table~\ref{tab:main} reports regret against the offline per-client Oracle over 594 configuration--$\lambda$ cells. The random and personalize-all policies are strongly suboptimal ($189.3\%\pm367.1\%$ and $409.0\%\pm803.2\%$ relative regret). The best-uniform policy---a single tuned depth for all clients---has only $6.5\%\pm7.8\%$ regret (CI $[5.9,7.1]$), despite an Oracle win rate of $82.5\%\pm38.0\%$ (CI $[79.5,85.4]$). Only 72.4\% of cells are Wilcoxon-significant versus the best-uniform policy (median $d_z{=}0.50$), and 18\% of cells favor that baseline. Per-client allocation can help, but the relevant comparator is the best-uniform policy---not only the global-only or personalize-all policies.

\begin{table}[t]
\caption{COST-AWARE REGRET VS.\ OFFLINE PER-CLIENT ORACLE ($n{=}594$). LOWER IS BETTER; 95\% BOOTSTRAP CIS.}
\label{tab:main}
\centering
\scriptsize
\resizebox{\columnwidth}{!}{%
\begin{tabular}{lccc}
\toprule
\textbf{Policy} & \textbf{Regret (\%)} & \textbf{Oracle win (\%)} & \textbf{95\% CI}\\
\midrule
Random & $189.3\pm367.1$ & $99.7\pm5.8$ & [160.6,219.8]\\
Best-uniform & $6.5\pm7.8$ & $82.5\pm38.0$ & [5.9,7.1]\\
Global-only & $28.2\pm29.6$ & $83.3\pm37.3$ & [25.9,30.6]\\
Personalize-all & $409.0\pm803.2$ & $99.2\pm9.1$ & [346.2,472.9]\\
\bottomrule
\end{tabular}%
}
\end{table}

\subsection{Finding 3: Full personalization becomes costly}
\begin{figure}[t]
\centering
\includegraphics[width=\linewidth]{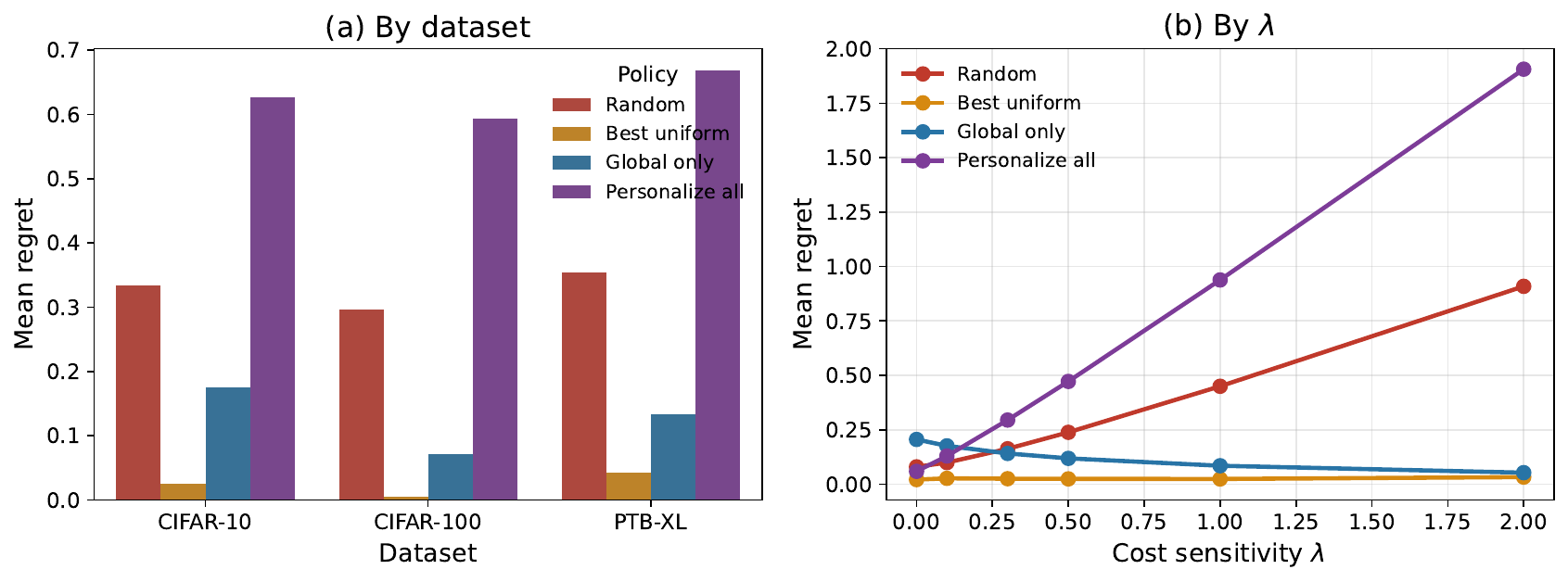}
\caption{Mean offline-Oracle regret of fixed policies by dataset and cost sensitivity $\lambda$.}
\label{fig:regret}
\end{figure}

Fig.~\ref{fig:phase}(d) and Fig.~\ref{fig:regret} show that full personalization becomes increasingly unattractive as $\lambda$ grows. The personalize-all policy is the worst fixed policy on average. At $\lambda{=}0$, deep actions can be useful; for $\lambda{>}0$, $P_5$ pays full-depth cost on every client, including those for which a shallow model already captures most predictive benefit.

Ditto reinforces this distinction without implying that Ditto is a poor accuracy optimizer: it performs full-model personalized optimization and is scored at P5-equivalent cost. On 486 configuration--$\lambda$ cells, Ditto regret is $454.9\%\pm875.8\%$ (CI $[380.7,534.7]$), compared with $6.4\%\pm8.0\%$ for the best-uniform policy and $27.9\%\pm29.7\%$ for the global-only policy (Table~\ref{tab:ditto}). Client-level checks confirm that Ditto normalized cost matches P5 (maximum absolute difference 0 across 81 configurations). When $\lambda{=}0$, Ditto still trails the Oracle and the best-uniform policy in Macro-F1 (relative regret 10.1\%, CI $[1.9,17.5]$), so the gap is not only a cost-accounting artifact.

\begin{table}[t]
\caption{DITTO UNDER THE SAME UTILITY (P5-EQUIVALENT COST).}
\label{tab:ditto}
\centering
\footnotesize
\begin{tabular}{lccc}
\toprule
\textbf{Policy} & \textbf{Regret (\%)} & \textbf{95\% CI} & \textbf{Oracle win (\%)}\\
\midrule
Ditto & $454.9\pm875.8$ & [380.7,534.7] & 94.0\\
Best-uniform & $6.4\pm8.0$ & [5.7,7.1] & 80.7\\
Global-only & $27.9\pm29.7$ & [25.3,30.6] & 81.7\\
\bottomrule
\end{tabular}
\end{table}

\subsection{Finding 4: Dataset modality matters}
Tables~\ref{tab:dataset}--\ref{tab:ptbxl} show that modality materially affects regret and Oracle assignment. CIFAR-100 is the strongest caution against overclaiming heterogeneous allocation: low average best-uniform regret ($5.5\%\pm8.5\%$) coincides with a relatively low Oracle win rate ($63.9\%\pm48.2\%$). The PTB-XL \emph{subset} (Lead~II, $\leq$5000 records, binary labels) retains both global and deep Oracle mass at $\lambda{=}0.3$, but this medical evidence is narrower than the CIFAR evaluation and should not be read as a full clinical deployment study.

\begin{table}[t]
\caption{BEST-UNIFORM REGRET VS.\ OFFLINE ORACLE BY DATASET.}
\label{tab:dataset}
\centering
\scriptsize
\begin{tabular}{lccc}
\toprule
\textbf{Dataset} & \textbf{Regret (\%)} & \textbf{Win (\%)} & \textbf{95\% CI}\\
\midrule
CIFAR-10 & $6.4\pm10.4$ & $81.7\pm38.8$ & [4.7,8.3]\\
CIFAR-100 & $5.5\pm8.5$ & $63.9\pm48.2$ & [4.4,6.8]\\
CIFAR-10 aux. & $7.1\pm6.7$ & $90.7\pm29.1$ & [5.9,8.4]\\
PTB-XL subset & $7.2\pm5.0$ & $96.7\pm18.0$ & [6.5,7.9]\\
\bottomrule
\end{tabular}
\end{table}

\begin{table}[t]
\caption{ORACLE SPARSITY AT $\lambda{=}0.3$ (SPARSITY-VALIDATION CELLS).}
\label{tab:sparsity}
\centering
\scriptsize
\resizebox{\columnwidth}{!}{%
\begin{tabular}{lcccc}
\toprule
\textbf{Dataset} & \textbf{P0 ratio} & \textbf{Deep ratio} & \textbf{Unif.\ gap} & \textbf{P5 gap}\\
\midrule
CIFAR-100 & $0.620\pm0.383$ & $0.006\pm0.016$ & $9.3\pm12.5$ & $136\pm32$\\
CIFAR-10 aux. & $0.326\pm0.297$ & $0.074\pm0.092$ & $7.1\pm4.4$ & $1850\pm3376$\\
PTB-XL & \multicolumn{4}{l}{see Table~\ref{tab:ptbxl}}\\
Overall & $0.499\pm0.317$ & $0.069\pm0.096$ & $9.1\pm8.3$ & $695\pm2082$\\
\bottomrule
\end{tabular}%
}
\end{table}

\begin{table}[t]
\caption{PTB-XL SUBSET OFFLINE-ORACLE BEHAVIOR (LEAD~II, $\leq$5000 RECORDS, BINARY).}
\label{tab:ptbxl}
\centering
\scriptsize
\begin{tabular}{lcc}
\toprule
\textbf{Metric} & \textbf{Mean $\pm$ std} & \textbf{95\% CI}\\
\midrule
P0 ratio ($\lambda{=}0.3$) & $0.552\pm0.167$ & [0.481,0.631]\\
Deep ratio ($\lambda{=}0.3$) & $0.127\pm0.110$ & [0.079,0.178]\\
Oracle vs uniform (rel.\ \%) & $11.0\pm5.6$ & [8.5,13.6]\\
Oracle vs P5-all (rel.\ \%) & $98.2\pm37.7$ & [82.2,116.0]\\
Regret vs uniform (all $\lambda$) & $7.2\pm5.0$ & [6.5,7.9]\\
\bottomrule
\end{tabular}
\end{table}

\section{Statistical Validation}
Wilcoxon summaries in Table~\ref{tab:sig} complement Table~\ref{tab:main}: 72.4\% of cells are significant versus the best-uniform policy, yet 18\% of cells favor that baseline, so significance is not universal. Oracle win rate (82.5\%) and regret (6.5\%) should be read jointly---a high win rate does not imply large utility gaps when the Oracle margin is small. The random and personalize-all policies are strongly dominated (median $d_z{=}1.07$ and $5.62$), supporting the cost-aware reading of full personalization.

\begin{table}[t]
\caption{WILCOXON SUMMARIES OVER 594 CELLS.}
\label{tab:sig}
\centering
\scriptsize
\resizebox{\columnwidth}{!}{%
\begin{tabular}{lcccc}
\toprule
\textbf{Policy} & \textbf{Sig.\ cells} & \textbf{Median $p$} & \textbf{Median $d_z$} & \textbf{Interp.}\\
\midrule
Random & 588/594 (99.0\%) & $9.8{\times}10^{-5}$ & 1.07 & Strong\\
Best-uniform & 430/594 (72.4\%) & 0.0017 & 0.50 & Small--mod.\\
Global-only & 428/594 (72.1\%) & 0.0011 & 0.66 & Moderate\\
Personalize-all & 580/594 (97.6\%) & $9.5{\times}10^{-7}$ & 5.62 & Strong (cost)\\
\bottomrule
\end{tabular}%
}
\end{table}

\section{Discussion and Implications}
\paragraph{Accuracy-only evaluation is incomplete.}
At $\lambda{=}0$, the offline Oracle selects deep personalization in isolated cells. Claims about sparsity or depth must specify $(U,\lambda)$; accuracy-only plots can overstate deep personalization and understate global-only efficiency when $\lambda{>}0$.

\paragraph{Best-uniform policy as deployment benchmark.}
As reported in Finding~2, the best-uniform policy is competitive across many configurations. Deployment systems should report regret against this policy before adopting heterogeneous allocation.

\paragraph{Objectives can diverge.}
Ditto results illustrate that accuracy-oriented and deployment-aware utilities may favor different operating points: full-model personalization is penalized under $U$ when marginal Macro-F1 gain does not offset full-depth cost.

\paragraph{Allocation maps as reporting practice.}
We advocate utility curves over $\lambda$, P0/deep ratios, and regret versus fixed policies with uncertainty bands---shifting from ``does personalization improve accuracy?'' to ``under which $(\text{task},\alpha,\lambda)$ is it worth its deployment cost?'' Table~\ref{tab:reporting} lists minimum reporting items.

\paragraph{Decision formulation vs.\ decision policy.}
The framework separates decision formulation from decision policy. This study addresses the former by defining candidate actions, the utility objective, and an offline Oracle reference, and by empirically characterizing the resulting decision landscape. It does not provide a deployment-time policy that selects an action without evaluating alternatives. The characterization may support subsequent lightweight rules, calibration procedures, learned policies, or adaptive schedulers, but such mechanisms remain outside the present scope.

\paragraph{Practical offline calibration.}
Exhaustive $P_0$--$P_5$ evaluation is intended for offline characterization rather than routine deployment. In practice, a system designer could apply the framework during a calibration or pilot phase on representative clients, identify which actions dominate under relevant heterogeneity and cost regimes, and use the resulting evidence to define a simpler deployment policy. Recalibration may be needed if the architecture, client population, hardware, cost model, or $\lambda$ range changes. Migrating from synthetic to real WFDB ECG lowered P0 ratio from 81.6\% to $0.552\pm0.167$ at $\lambda{=}0.3$, illustrating that modality choice can shift observed regimes.

\paragraph{Formulation-level generality vs.\ experiment-level scope.}
The same decision rule can, in principle, apply to other finite sets after redefining $\mathcal{A}$ and the cost model. Generality claimed here is methodological rather than empirical: experiments validate only the $P_0$--$P_5$ hierarchy.

\begin{table}[t]
\caption{RECOMMENDED REPORTING ITEMS FOR UTILITY-AWARE PFL EVALUATION.}
\label{tab:reporting}
\centering
\footnotesize
\begin{tabular}{ll}
\toprule
\textbf{Item} & \textbf{Purpose}\\
\midrule
Utility curve over $\lambda$ & Cost sensitivity\\
P0/deep allocation ratios & Regime patterns\\
Best-uniform baseline & Simple deployment test\\
Oracle regret + CIs & Headroom and stability\\
Cost/weight caveat & Bounds interpretation\\
\bottomrule
\end{tabular}
\end{table}

\section{Limitations}
\label{sec:limitations}
While the decision formulation accommodates arbitrary finite candidate sets, the present empirical findings are specific to the evaluated $P_0$--$P_5$ hierarchy, datasets, architectures, heterogeneity settings, cost normalization, default weights, and $\lambda$ range. We therefore do not claim that observed Oracle frequencies, regret values, or regime boundaries transfer unchanged to alternative PFL methods or deployment environments. The cost model is a normalized proxy rather than real-device measurement; $\lambda$ is a preference parameter, not a universally identifiable constant. The offline Oracle cannot be used directly during deployment. PTB-XL evidence is intentionally narrower than CIFAR (Lead~II subset, $\leq$5000 records, binary labels) and does not substitute for broader clinical evaluation. Additional limits: three seeds, simulated hardware multipliers, no native FEMNIST, one PFL comparator (Ditto), and uncorrected descriptive Wilcoxon tests. Future work includes alternative action spaces, real-device measurements, additional medical datasets, application-specific costs, and lightweight deployment policies derived from offline calibration.

\section{Conclusion}
Personalization value in federated learning is regime-dependent rather than universal.
Under $U=\mathrm{Performance}-\lambda\cdot\mathrm{Cost}$ with default weighting and $\mathcal{A}=\{P_0,\ldots,P_5\}$, offline Oracle depth varies with dataset, $\alpha$, and $\lambda$. High $\lambda$ concentrates mass on $P_0$ on several vision tasks; low $\lambda$ restores deeper personalization; modality shifts empirical regimes. Best-uniform remains close ($6.5\%\pm7.8\%$; CI $[5.9,7.1]$; 82.5\% win rate); personalize-all and random do not. At $\lambda{=}0.3$, mean P0 ratio $0.499\pm0.317$ rejects universal global-only and universal deep-personalization laws. This work does not propose a deployment-time personalization algorithm. Instead, it formulates personalization-depth selection as a utility-based empirical decision problem and characterizes its offline decision landscape using an Oracle reference. The resulting evidence can inform future calibration rules or lightweight deployment policies, whose design and validation remain outside the present scope, including alternative PFL methods, parameter-efficient adaptation, real-device measurements, additional medical datasets, and application-specific cost functions.

\section*{Impact Statement}
This work analyzes resource--accuracy trade-offs using public CIFAR and PTB-XL releases; all datasets used in this study are publicly available, and no new patient-identifiable data are collected. Source code is available at \url{https://github.com/bbyuksel}.

\vspace{-6pt}
\balance
{\fontsize{6}{6.75}\selectfont
\setlength{\itemsep}{0pt}%
\setlength{\parsep}{0pt}%
\setlength{\parskip}{0pt}%

}

\begin{thebibliography}{44}

\bibitem{mcmahan2017fedavg}
H.~B. McMahan, E.~Moore, D.~Ramage, S.~Hampson, and B.~A. y Arcas,
``Communication-efficient learning of deep networks from decentralized data,''
in \emph{Proc. AISTATS}, 2017, pp. 1273--1282.

\bibitem{li2020fedprox}
T.~Li et al.,
``Federated optimization in heterogeneous networks,''
\emph{Proc. Mach. Learn. Syst.}, vol.~2, pp. 429--450, 2020.

\bibitem{karimireddy2020scaffold}
S.~P. Karimireddy et al.,
``SCAFFOLD: stochastic controlled averaging for federated learning,''
in \emph{Proc. ICML}, 2020, pp. 5132--5143.

\bibitem{wang2020fednova}
J.~Wang, Q.~Liu, H.~Liang, G.~Joshi, and H.~V. Poor,
``Tackling the objective inconsistency problem in heterogeneous federated optimization,''
\emph{Adv. Neural Inf. Process. Syst.}, vol.~33, pp. 7611--7623, 2020.

\bibitem{reddi2021adaptive}
S.~J. Reddi et al.,
``Adaptive federated optimization,''
in \emph{Proc. ICLR}, 2021.

\bibitem{kairouz2021advances}
P.~Kairouz and H.~B. McMahan,
``Advances and open problems in federated learning,''
\emph{Found. Trends Mach. Learn.}, vol.~14, no.~1--2, pp. 1--210, 2021.

\bibitem{bonawitz2019towards}
K.~Bonawitz et al.,
``Towards federated learning at scale: system design,''
\emph{Proc. Mach. Learn. Syst.}, vol.~1, pp. 374--388, 2019.

\bibitem{hsieh2020noniid}
T.-M. H. Hsu, H.~Qi, and M.~Brown,
``Measuring the effects of non-identical data distribution for federated visual classification,''
unpublished.

\bibitem{zhao2018noniid}
Y.~Zhao et al.,
``Federated learning with non-IID data,''
unpublished.

\bibitem{fallah2020perfedavg}
A.~Fallah, A.~Mokhtari, and A.~Ozdaglar,
``Personalized federated learning with theoretical guarantees: a model-agnostic meta-learning approach,''
\emph{Adv. Neural Inf. Process. Syst.}, vol.~33, pp. 3557--3568, 2020.

\bibitem{li2021ditto}
T.~Li, S.~Hu, A.~Beirami, and V.~Smith,
``Ditto: fair and robust federated learning through personalization,''
in \emph{Proc. ICML}, 2021, pp. 6357--6368.

\bibitem{dinh2020pfedme}
C.~T. Dinh, N.~Tran, and J.~Nguyen,
``Personalized federated learning with Moreau envelopes,''
\emph{Adv. Neural Inf. Process. Syst.}, vol.~33, pp. 21394--21405, 2020.

\bibitem{arivazhagan2019fedper}
M.~G. Arivazhagan, V.~Aggarwal, A.~K. Singh, and S.~Choudhary,
``Federated learning with personalization layers,''
unpublished.

\bibitem{collins2021fedrep}
L.~Collins, H.~Hassani, A.~Mokhtari, and S.~Shakkottai,
``Exploiting shared representations for personalized federated learning,''
in \emph{Proc. ICML}, 2021, pp. 2089--2099.

\bibitem{oh2022fedbabu}
J.~Oh, S.~Kim, and S.-Y. Yun,
``FedBABU: toward enhanced representation for federated image classification,''
in \emph{Proc. ICLR}, 2022.

\bibitem{deng2020apfl}
Y.~Deng, M.~M. Kamani, and M.~Mahdavi,
``Adaptive personalized federated learning,''
unpublished.

\bibitem{huang2021fedamp}
Y.~Huang et al.,
``Personalized cross-silo federated learning on non-IID data,''
in \emph{Proc. AAAI}, vol.~35, no.~9, 2021, pp. 7865--7873.

\bibitem{li2021fedbn}
X.~Li, M.~Jiang, X.~Zhang, M.~Kamp, and Q.~Dou,
``FedBN: federated learning on non-IID features via local batch normalization,''
in \emph{Proc. ICLR}, 2021.

\bibitem{zhang2023pfedlora}
L.~Yi, H.~Yu, G.~Wang, X.~Liu, and X.~Li,
``pFedLoRA: model-heterogeneous personalized federated learning with LoRA tuning,''
unpublished.

\bibitem{hu2021lora}
E.~J. Hu et al.,
``LoRA: low-rank adaptation of large language models,''
in \emph{Proc. ICLR}, 2022.

\bibitem{zhang2023fedala}
J.~Zhang et al.,
``FedALA: adaptive local aggregation for personalized federated learning,''
in \emph{Proc. AAAI}, vol.~37, no.~9, 2023, pp. 11237--11244.

\bibitem{zhu2024fedtsd}
H.~Zhu, Y.~Fan, and Z.~Xie,
``Federated two-stage decoupling with adaptive personalization layers,''
\emph{Complex Intell. Syst.}, vol.~10, no.~3, pp. 3657--3671, 2024.

\bibitem{xing2025fedfsa}
X.~Xing et al.,
``Flexible sharpness-aware personalized federated learning,''
in \emph{Proc. AAAI}, vol.~39, no.~20, 2025, pp. 21707--21715.

\bibitem{wu2025pfedcs}
S.~Wu et al.,
``PFedCS: a personalized federated learning method for enhancing collaboration among similar classifiers,''
in \emph{Proc. AAAI}, vol.~39, no.~20, 2025, pp. 21572--21580.

\bibitem{wang2024flora}
Z.~Wang et al.,
``FLoRA: federated fine-tuning large language models with heterogeneous low-rank adaptations,''
in \emph{Proc. NeurIPS}, 2024.

\bibitem{ye2024openfedllm}
R.~Ye et al.,
``OpenFedLLM: training large language models on decentralized private data via federated learning,''
in \emph{Proc. ACM KDD}, 2024, pp. 6137--6147.

\bibitem{stich2018localsgd}
S.~U. Stich,
``Local SGD converges fast and communicates little,''
in \emph{Proc. ICLR}, 2019.

\bibitem{alistarh2017qsgd}
D.~Alistarh, D.~Grubic, J.~Li, R.~Tomioka, and M.~Vojnovic,
``QSGD: communication-efficient SGD via gradient quantization and encoding,''
\emph{Adv. Neural Inf. Process. Syst.}, vol.~30, 2017.

\bibitem{konevcny2016federated}
J.~Kone\v{c}n{\'y} et al.,
``Federated learning: strategies for improving communication efficiency,''
unpublished.

\bibitem{nishio2019client}
T.~Nishio and R.~Yonetani,
``Client selection for federated learning with heterogeneous resources in mobile edge,''
in \emph{Proc. IEEE ICC}, 2019, pp. 1--7.

\bibitem{lai2021oort}
F.~Lai, X.~Zhu, H.~V. Madhyastha, and M.~Chowdhury,
``Oort: efficient federated learning via guided participant selection,''
in \emph{Proc. USENIX OSDI}, 2021, pp. 19--35.

\bibitem{cho2020tifl}
Z.~Chai et al.,
``TiFL: a tier-based federated learning system,''
in \emph{Proc. ACM HPDC}, 2020, pp. 125--136.

\bibitem{diao2021heterofl}
E.~Diao, J.~Ding, and V.~Tarokh,
``HeteroFL: computation and communication efficient federated learning for heterogeneous clients,''
in \emph{Proc. ICLR}, 2021.

\bibitem{wang2023personalized}
Z.~Wang, X.~Ma, H.~Zhang, and D.~Yuan,
``Communication-efficient personalized federated learning for digital twin in heterogeneous industrial IoT,''
in \emph{Proc. IEEE ICC Workshops}, 2023, pp. 237--241.

\bibitem{tan2022surveypfl}
A.~Z. Tan, H.~Yu, L.~Cui, and Q.~Yang,
``Towards personalized federated learning,''
\emph{IEEE Trans. Neural Netw. Learn. Syst.}, vol.~34, no.~12, pp. 9587--9603, 2023.

\bibitem{caldas2018leaf}
S.~Caldas et al.,
``LEAF: a benchmark for federated settings,''
unpublished.

\bibitem{lai2022fedscale}
F.~Lai et al.,
``FedScale: benchmarking model and system performance of federated learning at scale,''
in \emph{Proc. ICML}, 2022, pp. 11814--11827.

\bibitem{he2020fedml}
C.~He et al.,
``FedML: a research library and benchmark for federated machine learning,''
unpublished.

\bibitem{ogierdu2022flamby}
J.~Ogier du Terrail et al.,
``FLamby: datasets and benchmarks for cross-silo federated learning in realistic healthcare settings,''
\emph{Adv. Neural Inf. Process. Syst.}, vol.~35, pp. 5315--5334, 2022.

\bibitem{wagner2020ptbxl}
P.~Wagner et al.,
``PTB-XL, a large publicly available electrocardiography dataset,''
\emph{Sci. Data}, vol.~7, no.~1, p.~154, 2020.

\bibitem{yuksel2026ecgfl}
B.~B. Yuksel and A.~Y. Metin,
``Federated learning with homomorphic encryption for secure ECG anomaly detection,''
\emph{Biomed. Signal Process. Control}, vol.~116, Art.\ no.~109557, 2026.

\bibitem{yuksel2026hardwarefl}
B.~B. Y\"uksel and E.~Dikbiyik,
``Hardware-aware federated learning for speech emotion recognition,''
in \emph{Proc. 34th Signal Process. Commun. Appl. Conf. (SIU)}, Jul. 2026, pp. 1--4.

\end{thebibliography}
\end{document}